\documentclass[10pt,twocolumn,letterpaper]{article}

\usepackage{wacv}
\usepackage{times}
\usepackage{epsfig}
\usepackage{graphicx}
\usepackage{amsmath}
\usepackage{amssymb}
\usepackage{booktabs}
\usepackage{xcolor}
\def\wacvPaperID{745} % Enter the WACV Paper ID here

\wacvfinalcopy % *** Uncomment this line for the final submission

\ifwacvfinal
\usepackage[breaklinks=true,bookmarks=false]{hyperref}
\else
\usepackage[pagebackref=true,breaklinks=true,colorlinks,bookmarks=false]{hyperref}
\fi

\usepackage{paralist}

\begin{document}

%%%%%%%%% TITLE
\title{Exploring Transformer Backbones for Image Diffusion Models}

\author{Princy Chahal\\
George Brown College\\
% For a paper whose authors are all at the same institution,
% omit the following lines up until the closing ``}''.
% Additional authors and addresses can be added with ``\and'',
% just like the second author.
% To save space, use either the email address or home page, not both
}

\maketitle
\thispagestyle{empty}

\begin{abstract}
   \noindent We present an end-to-end Transformer based Latent Diffusion model for image synthesis. On the ImageNet class conditioned generation task we show that a Transformer based Latent Diffusion model achieves a 14.1FID which is comparable to the 13.1FID score of a UNet based architecture. In addition to showing the application of Transformer models for Diffusion based image synthesis this simplification in architecture allows easy fusion and modeling of text and image data. The multi-head attention mechanism of Transformers enables simplified interaction between the image and text features which removes the requirement for cross-attention mechanism in UNet based Diffusion models.\\
\end{abstract}
\section{Introduction}

The vision community has witnessed a renewed interest in usage of Transformers ~\cite{DBLP:journals/corr/VaswaniSPUJGKP17} for computer vision tasks, ~\cite{DBLP:conf/iclr/DosovitskiyB0WZ21} introduced ViT (Vision Transformers) which showed competitive performance compared to CNNs on standard vision tasks (e.g. ImageNet classification),~\cite{bao2022beit, DBLP:journals/corr/abs-2111-06377} applied ViT to the task of self-supervised learning, the architecture showed performance better than CNNs on this task. In this work we intend to explore the usage of Transformer models for image synthesis using Diffusion Probabilistic Models ~\cite{DBLP:conf/nips/HoJA20}. Our experiments on the ImageNet dataset show that Transformers are a competitive replacement for CNN based UNet architectures.

The previous few years have seen the rise of diffusion based models for image generation, ~\cite{DBLP:journals/corr/abs-2204-06125, DBLP:journals/corr/abs-2205-11487} showed photo-realistic image generation with diffusion model. Diffusion models use a CNN UNet based architecture to process images. The success of Transformer models in NLP has paved the way for application of multi-head attention based architectures for Vision models. Taking inspiration from application of transformers to NLP, image classification and detection tasks we explore the usage of an end-to-end transformer system for diffusion based image generation. We build on top of ~\cite{DBLP:journals/corr/abs-2112-10752} which performs latent diffusion by first encoding the images to a latent space and then applying a UNet ~\cite{DBLP:journals/corr/unet} architecture to remove noise in the diffusion, as the final step then the latent space is decoded to the original image space. We show that on class conditioned ImageNet dataset generation task we achieve a FID of \textbf{14.1} which is competitive compared to the FID of \textbf{13.1} of a UNet based latent diffusion model (lower FID is better).

\begin{figure*}[ht]
\begin{center}
    \makebox[\textwidth]{\includegraphics[width=\textwidth]{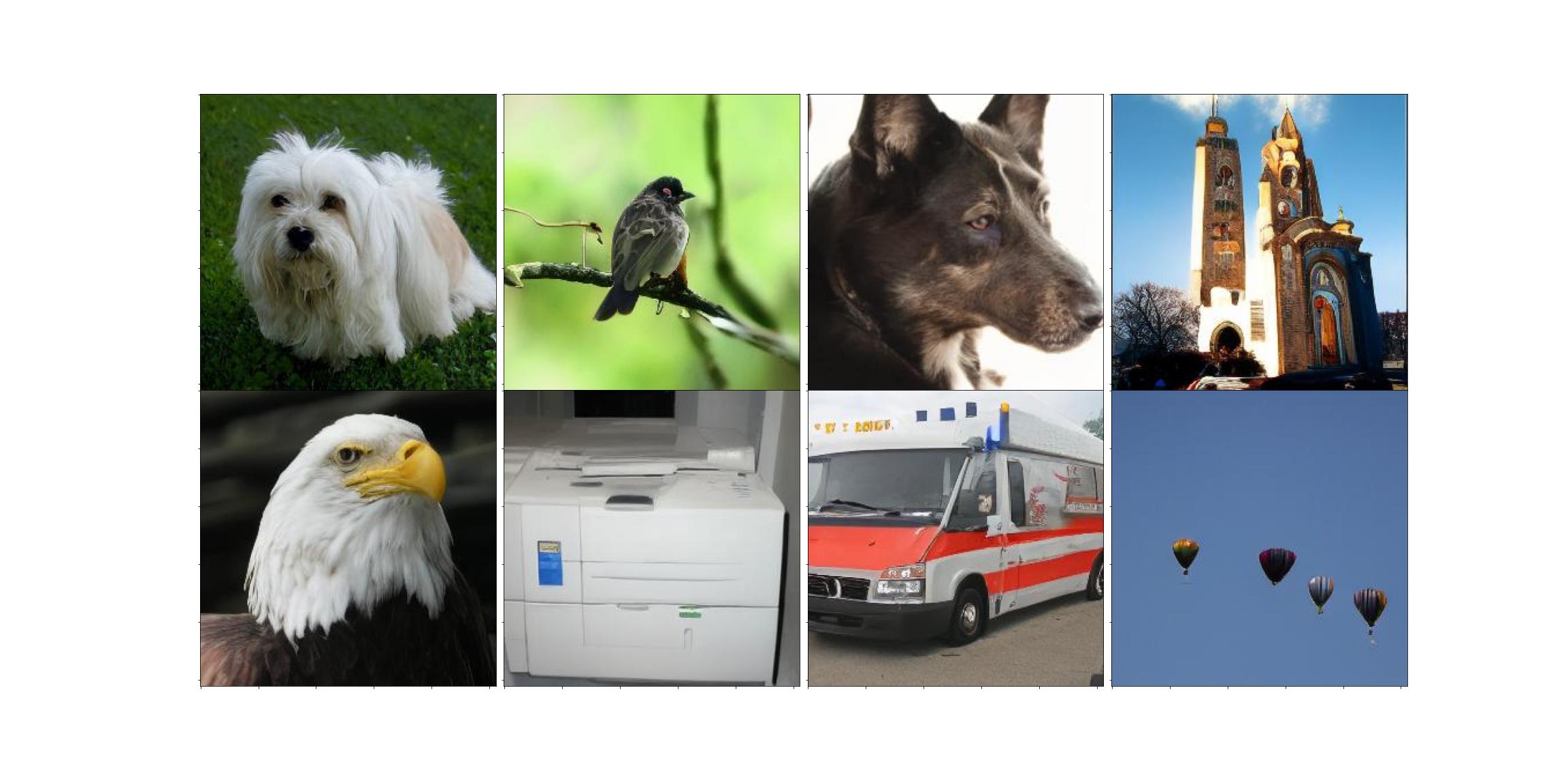}}
    \caption{Class conditioned sample images from Transformer and UNet models. The class labels for images shown in the figure are \{\textit{Lhasa, Bulbul, Kelpie, Church, Bald Eagle, Photocopier, Ambulance, Balloon}\}. The samples shown in the image were taken from our best Transformer model \textbf{LDM-8-G-Trainsformer (14.1FID)} from table \ref{table:fid_results}}.
    \label{image_samples}
\end{center}
\end{figure*}

This work is another step in the direction of usage of the Transformer architecture for a vision task. We hope that this work enables easy modeling of text and image for diffusion based image generation as a uniform architecture is used across the two different modalities. Specifically our work makes the following contributions:
\begin{asparaenum}
    \item We propose a Transformer architecture for performing latent diffusion, the proposed architecture simplifies the interaction between text and image modalities for diffusion based image generation.
    \item We benchmark the performance of Transformer and UNet based architectures on the task of class conditioned image generation. Our experiments show that the Transformer architecture performs on-par with the Convolutional UNet based architecture.
\end{asparaenum}

\hfill

\section{Related Work}
Generative models have seen tremendous interest from the computer vision community in the past few years, ~\cite{DBLP:conf/nips/GoodfellowPMXWOCB14} introduced Generative Adversarial Networks (GANs) which pose the image generation task as a minimax game between a generator and a discriminator. GANs allow efficient sampling of high resolution images but the framework has been known to cause optimization difficulties ~\cite{DBLP:journals/corr/abs-1801-04406}. Another generative modeling framework is the Variational Autoencoder ~\cite{DBLP:journals/corr/KingmaW13} line of work. Variational Autoencoder enable efficient synthesis of images compared to other generative methods but the samples produced by Variational Autoencoders are of lower quality compared to GANs.

Diffusion probabilistic models for image generation were introduced in ~\cite{DBLP:conf/nips/HoJA20}, in Diffusion models the task of image generation is formulated as a denoising process. Until now, UNet architecture ~\cite{DBLP:conf/nips/HoJA20, DBLP:journals/corr/abs-2205-11487, DBLP:journals/corr/abs-2112-10752} has been an important component of the Diffusion models. Recent works ~\cite{DBLP:journals/corr/abs-2205-11487, DBLP:journals/corr/abs-2112-10752, DBLP:journals/corr/abs-2204-06125} have shown state of the art performance on class and text conditioned image generation with diffusion models, these works use a UNet architecture for networks in the diffusion process. Latent diffusion models ~\cite{DBLP:journals/corr/abs-2112-10752} improved the efficiency and reduced the resources required for diffusion models by using VQGAN ~\cite{DBLP:journals/corr/abs-2012-09841} encoder for mapping the image from the pixel space ($256 \times 256$ RGB) to the latent space ($32 \times 32$). After the encoding the diffusion process is carried out in the latent space and as the last step a VQGAN decoder is used to map the latent space back to the image space.

UNet, a convolutional neural net (CNN~\cite{DBLP:journals/pieee/LeCunBBH98}) based architecture was introduced by ~\cite{DBLP:journals/corr/unet} for biomedical image segmentation, since its introduction the UNet architecture has also found use in image synthesis diffusion work with ~\cite{DBLP:journals/corr/abs-2204-06125,DBLP:journals/corr/abs-2112-10752} using the architecture in their diffusion modules. In recent years Transformer architecture ~\cite{DBLP:journals/corr/VaswaniSPUJGKP17} has found wide application in NLP on machine translation ~\cite{DBLP:journals/corr/VaswaniSPUJGKP17} and language modeling tasks, for example BERT ~\cite{DBLP:conf/naacl/DevlinCLT19}, GPT ~\cite{DBLP:journals/corr/abs-2005-14165}. Vision Transformer (ViT) introduced in ~\cite{DBLP:conf/iclr/DosovitskiyB0WZ21} show competitive performance compared to CNNs on the task of image classification. In-line with this observation Transformers have also shown performance comparable to CNNs on object detection \cite{DBLP:conf/eccv/CarionMSUKZ20, DBLP:conf/iclr/ZhuSLLWD21} task. Recent works show that the Transformer architecture is relevant in the computer vision domain and we wanted to further explore its application for the task of diffusion based generative models. Transformers have been used for image generation in the past, \cite{DBLP:conf/icml/ChenRC0JLS20, DBLP:journals/corr/abs-2012-09841} in an autoregressive manner. In our work we explore application of Transformers for diffusion based image generation.
\section{Latent Diffusion Transformer}

For our Latent Diffusion Transformer we build on top of the Latent Diffusion framework proposed in ~\cite{DBLP:journals/corr/abs-2112-10752}. In this section we will introduce our Latent Diffusion Transformer architecture. We will briefly discuss the process of Diffusion for image generation and the Latent Diffusion framework, finally we will go through how the Vision Transformer (ViT) model fits into the Latent Diffusion framework.

\subsection{Diffusion Probabilistic Models}

Denoising Diffusion models ~\cite{DBLP:conf/nips/HoJA20} can be described as probabilistic models which denoise a Gaussian distributed random variable, in the context of image generation these models denoise a Gaussian image over $T$ timesteps. Diffusion denoising consists of two processes, forward diffusion process which is fixed to a Markov chain and a Gaussian noise is added to the sample at each time-step. In the forward diffusion process we start with a RGB image $x_0$ and gradually add noise to it until we reach the $T$ time-step and we get a noisy image $x_T$, forward diffusion: $x_0 \rightarrow x_1 \rightarrow \ldots \rightarrow x_T$. In the reverse diffusion process we remove noise from the noisy image $x_t \rightarrow x_{t-1}$ and try to obtain the original image $x_0$. For simplicity we can interpret these models as denoising autoencoders $\epsilon_\theta(x_t, t)$. The objective function for training models in the reverse diffusion process can be written as:

\begin{equation}
    L_{DM} = \mathbb{E}_{\varepsilon(x), \epsilon \sim \mathcal{N}(0,1),t} \left[||\epsilon - \epsilon_\theta(x_t, t)||^2_2 \right]
\end{equation}

In the reverse diffusion process $\epsilon_\theta(x_t, t)$ is modeled by a neural network, ~\cite{DBLP:conf/nips/HoJA20, DBLP:journals/corr/abs-2205-11487, DBLP:journals/corr/abs-2112-10752} modeled this using a CNN based time conditional UNet architecture ~\cite{DBLP:journals/corr/unet}. The diffusion process can be used to generate unconditional as well as class conditioned image generation. For our work we focus on class conditioned image generation, under this setting an addition class label $y$ variable will be added to our training objective function, the updated objective function can be written as:

\begin{equation}
    L_{DM} = \mathbb{E}_{\varepsilon(x), y, \epsilon \sim \mathcal{N}(0,1),t} \left[||\epsilon - \epsilon_\theta(x_t, t, \tau_\theta(y))||^2_2 \right]
\end{equation}

\noindent where $\tau_\theta(y)$ models the conditioning of the class variable. UNet and Transformer model $\tau_\theta$ differently which we will discuss in the subsequent sections.

\subsection{Latent Diffusion}

Image generation using Diffusion Probabilistic Models is a resource intensive method, for example ~\cite{DBLP:journals/corr/abs-2105-05233} reports using 150-1000 V100 GPU days for training models on the ImageNet dataset ~\cite{DBLP:conf/cvpr/DengDSLL009}. Latent Diffusion Models ~\cite{DBLP:journals/corr/abs-2112-10752} propose a shift from the pixel space $(256 \times 256)$ to a latent space $(32 \times 32)$. The pixel space is encoded/decoded to a latent space using the VQGAN encoder/decoder ~\cite{DBLP:journals/corr/abs-2012-09841} respectively. For the latent space: $z_t = \text{Encoder}_{VQGAN}(x_t)$, the training objective can be written as:

\begin{equation}
    L_{DM} = \mathbb{E}_{\varepsilon(x), y, \epsilon \sim \mathcal{N}(0,1),t} \left[||\epsilon - \epsilon_\theta(z_t, t, \tau_\theta(y))||^2_2 \right]
\end{equation}

For modeling the class conditional aspects of image generation Latent Diffusion Models use a cross-attention mechanism between the class label $+$ timestep embedding with features from the UNet architecture. Compared to Transformers which natively work on sequences this type of modeling is custom for UNet architectures, and is something we aim to simplify with the introduction of Transformer architecture in the Latent Diffusion process.

The diffusion process and latent space usage allow the model to efficiently generate images compared to the diffusion models ~\cite{DBLP:journals/corr/abs-2105-05233} and other autoregressive methods ~\cite{DBLP:journals/corr/abs-2112-10752, DBLP:conf/icml/ChenRC0JLS20}. Latent diffusion models show competitive state-of-the-art performance compared to ~\cite{DBLP:journals/corr/abs-2105-05233} on standard datasets like ImageNet.

\begin{figure}[ht]
\centering
    \includegraphics[width=\linewidth]{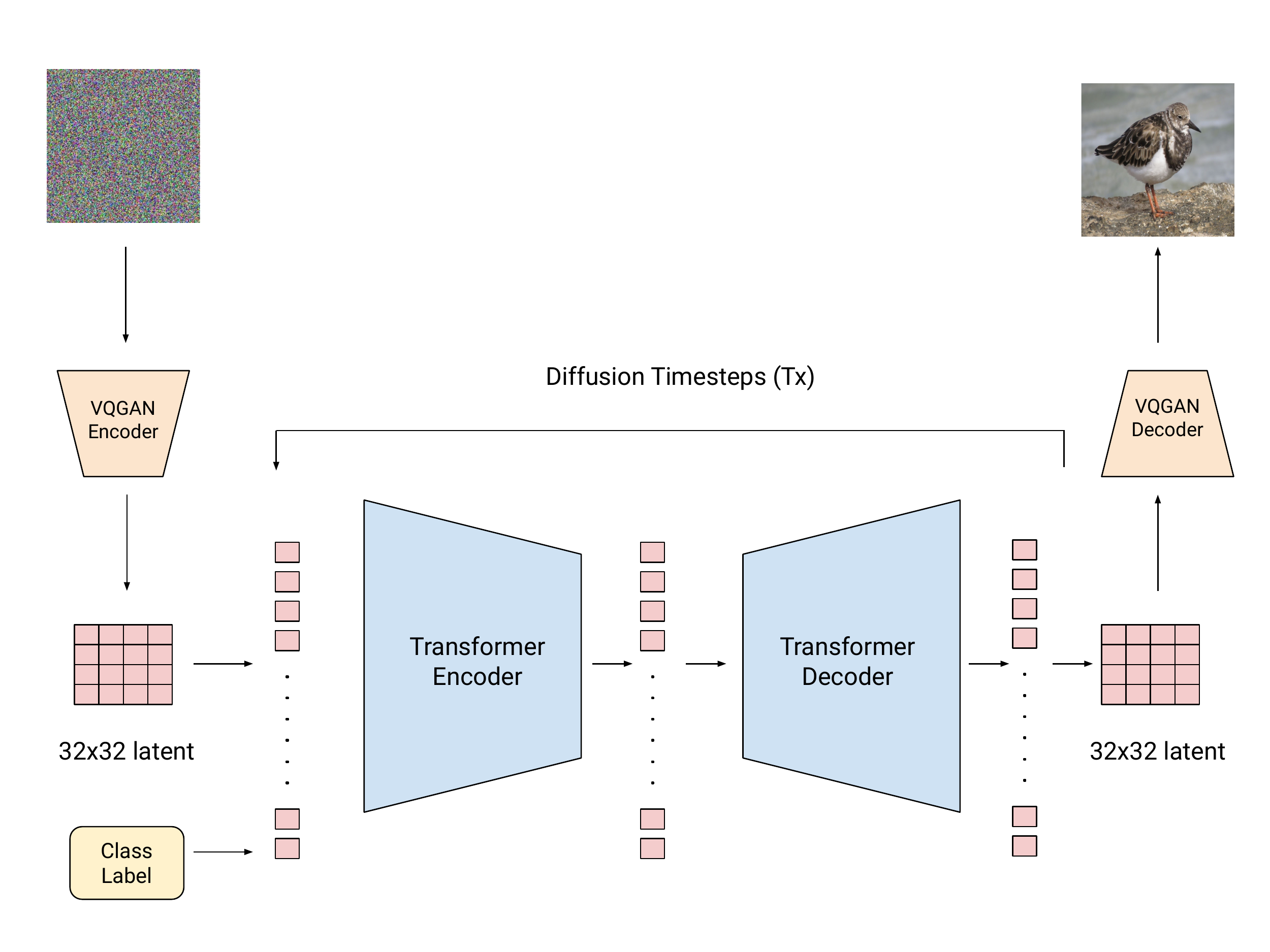}
    \caption{Latent Diffusion Transformer denoising (reverse-diffusion) process. The denoising process is repeated for $Tx$ timesteps, we use 200 timesteps for our experiments. A resolution $256 \times 256$ is used for the pixel space, and a resolution of $32 \times 32$ is used for the latent space. The denoised latent space vector generated by the Transformer is decoded to the RGB space using the VQGAN decoder.}
    \label{latent_diffusion_transformer_pipeline}
\end{figure}

\subsection{Latent Diffusion Transformer}

\noindent\textbf{Vision Transformer (ViT)}

\noindent Transformer ~\cite{DBLP:journals/corr/VaswaniSPUJGKP17} were traditionally used in NLP for machine translation, natural generation. They were introduced to get around the inefficiency of autoregressive LSTM based NLP models. ViT was introduced by ~\cite{DBLP:conf/iclr/DosovitskiyB0WZ21}, this model patchifies images into a series of patches which can then be treated as tokens to be passed through a Transformer model. Patches are passed through a linear embedding after which position embeddings are added into the patch embeddings to encode the location information of patches. To perform classification, a learnable ``Classification Token'' is added to the sequence of patches, we use a similar token for the class label based image generation.

Building on top of ViT`'s encoder architecture Masked Autoencoders ~\cite{DBLP:journals/corr/abs-2111-06377} introduced a ViT based encoder-decoder architecture for learning image embeddings in a self-supervised way. We use a similar architecture for our Latent Diffusion Transformer. For a fair comparison we use approximately same number of parameters and flops for our architecture related to the UNet architecture, for more details refer to the Table \ref{table:modelcompare}.

\hfill

\begin{table}[h!]
  \begin{center}
    \label{tab:table1}
    \begin{tabular}{l|r}
      \toprule % <-- Toprule here
      \textbf{Model Architecture} & \textbf{No. of params}\\
      \midrule % <-- Midrule here
      LDM-8-UNet & 395M\\
      LDM-8-Transformer & 310M\\
      \bottomrule % <-- Bottomrule here
    \end{tabular}
    % \addlinespace
    % \addlinespace
    \caption{Comparison for UNet \& Transformer architectures for Latent Diffusion, the architectures listed here were used for class conditional generation on ImageNet dataset. We report the number of parameters and count of Floating Point Operations per Second. (FLOPS).}\label{table:modelcompare}
  \end{center}
\end{table}

\noindent\textbf{ViT Encoder and Decoder}

\begin{figure*}[ht]
\begin{center}
    \makebox[\textwidth]{\includegraphics[width=\textwidth]{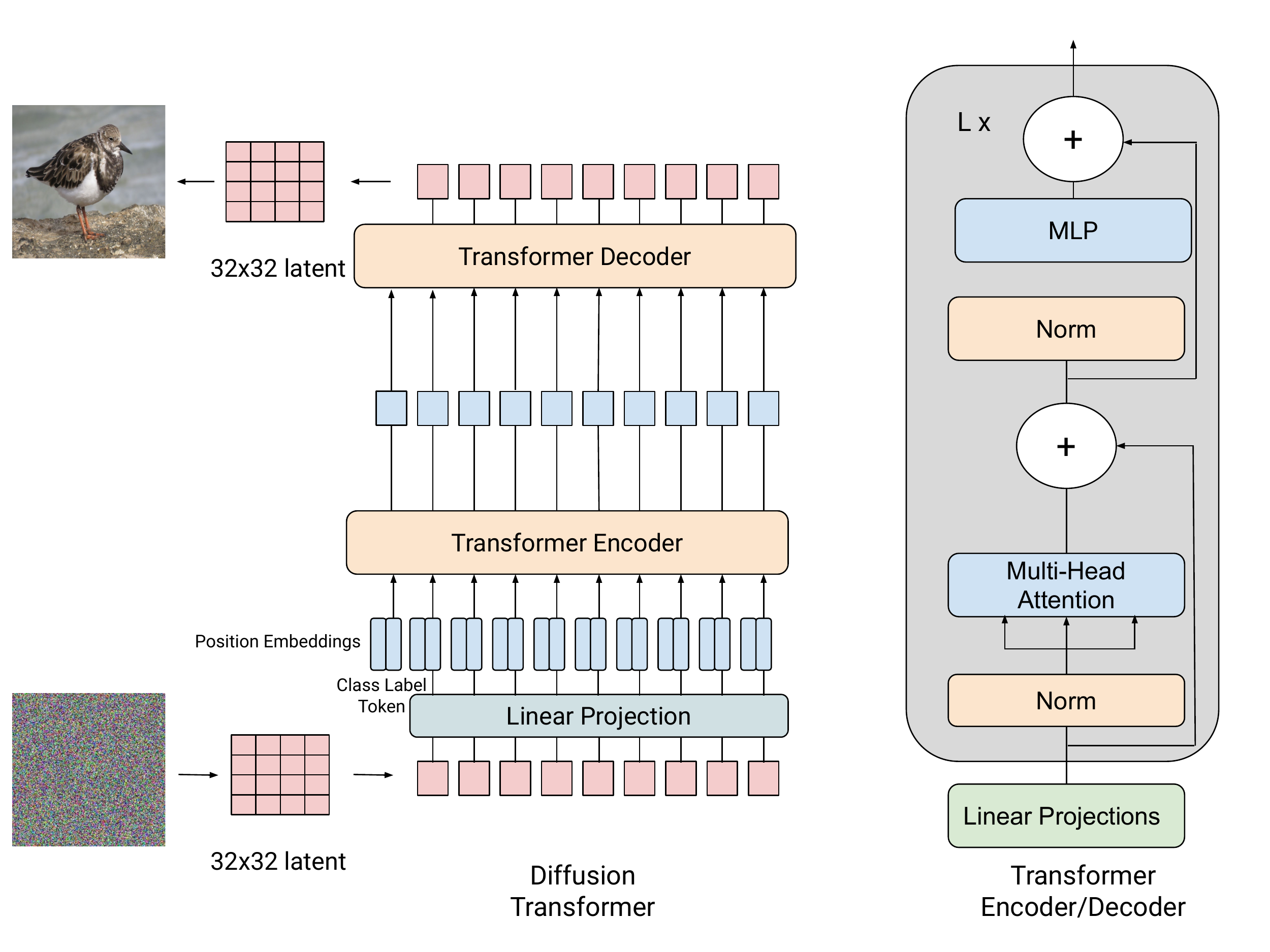}}
    \caption{Latent Diffusion Transformer architecture: we borrow the basic encoder, decoder architecture from ~\cite{DBLP:conf/iclr/DosovitskiyB0WZ21}. We use the same architecture for Encoder and Decoder with 12 layers for both the modules. The class label token is appended to the tokens before they are fed into the Encoder. The position embeddings consists of sinusoidal position embeddings for patch location in image and also the timestep embeddings for the timestep of the diffusion process.}
    \label{transformer_architecture}
\end{center}
\end{figure*}

\noindent For both the encoder and decoder we use the ViT architecture. The input latent space to encoder is of shape $32 \times 32$, we use a patch size of 2 on the space as a result of which we get $16 \times 16$, $256$ tokens which are then fed as input to the Transformer. For both the encoder and decoder we use 16 blocks of ViT. For a visualization and more details of the Transformer architecture refer to fig ~\ref{transformer_architecture}.

\hfill

\noindent\textbf{Position embeddings}: For retaining the positional information of patches positional embeddings are added to the patch embeddings. We use the sine-cosine version of positional embeddings ~\cite{DBLP:journals/corr/VaswaniSPUJGKP17} for our experiments. These embeddings are used in conjunction with the timestep embedding and are added to patch embeddings before being fed to the Transformer Encoder (refer to fig \ref{transformer_architecture}).

\hfill

\noindent\textbf{Class label token}: To model the label for class conditioned image generation we use a separate token similar to ~\cite{DBLP:conf/iclr/DosovitskiyB0WZ21}. For this token we linearly project the class label to an embedding space which is then concatenated with embeddings for the other $16 \times 16=256$ tokens.

\hfill

\noindent\textbf{Timestep embeddings}: To model the timestep $t$ for the denoising reverse diffusion process ($x_T \rightarrow x_{T-1} \rightarrow \ldots \rightarrow x_0$) we use sinusoidal embeddings. The same timestep embedding is added to embeddings of different patches and the class token.

\hfill

\hfill

\noindent With the above components we don't need to specifically model the interaction between the class label and image features using a custom cross-attention module. Usage of a native Transformer pipeline allows us to model the interaction between patches of latent and class label using the attention mechanism.

\hfill

\noindent\textbf{Classifier Free Guidance}

\noindent Following the results of ~\cite{ho2021classifierfree} we use a classifier free guidance for improving the quality of generated images. To train the classifier free label embedding we dropout the image labels for a random subset of 10\% image, label instances. At inference time we generate two samples, class conditioned and class free. Following ~\cite{ho2021classifierfree} the two samples are interpolated with a guidance scale of 1.25. Classifier free guidance significantly improves the quality and FID metric of our class conditioned generated images, this improvement is consistent across our UNet experiments and Transformer experiments.
\section{Experiments}

\noindent In this section we will discuss the experiment settings and results for our Latent Diffusion Transformer setup. We compare the proposed architecture against the standard UNet architecture on the ImageNet task ~\cite{DBLP:conf/cvpr/DengDSLL009}. For our experiments we report numbers only on the ImageNet dataset as this dataset is considered representative to measure quality of image synthesis models. For our experiments we built on top of the Latent Diffusion Models codebase ~\cite{DBLP:journals/corr/abs-2112-10752} as we were able to reproduce the numbers reported in the paper with the codebase.

\subsection{Model Architecture and Hyperparameter}

\noindent For our Transformer encoder-decoder architecture we use the standard ViT model with 12 layers each for the encoder and decoder. We use a similarly sized UNet model as is shown in Table \ref{table:modelcompare}. For both the architectures we use the same training loss and same hyperparameters for components that are shared across the architectures like number of diffusion steps, training dataset iterations, size of timestep embedding. We use the encoder decoders from the LDM-8 variant of ~\cite{DBLP:journals/corr/abs-2112-10752} for both the UNet and Transformer experiments as at the time it was the best reported VQGAN encoder-decoder model in ~\cite{DBLP:journals/corr/abs-2112-10752}.

\hfill

\subsection{Dataset}

\noindent For our training and validation we use the train and val splits from the ImageNet dataset ~\cite{DBLP:conf/cvpr/DengDSLL009}. For our initial experiments and hyperparameters we used smaller datasets like CIFAR100, LSUN. The numbers in table \ref{table:fid_results} are reported on the ImageNet dataset as that is the largest and most diverse datasets that the community uses for benchmarking generative models. The ImageNet dataset consists of $\sim 1$ Million training instances and we use the train split for learning the model and val split for measuring the FID performance.

\hfill

\subsection{FID Metric}

\noindent To measure the sample quality of our generative models we use the ``Frechet Inception Distance (FID)'' metric ~\cite{DBLP:conf/nips/HeuselRUNH17}. We calculate FID by generating 50k images and using the ImageNet val dataset as reference for the calculation. For more details on how we calculate FID please refer to the supplemental.

\hfill

\subsection{Training Details}

\noindent To train the Latent Diffusion models we use Tesla K80 nodes. We use the pre-trained frozen VQGAN encoder decoder models from ~\cite{DBLP:conf/nips/HeuselRUNH17} which were trained in an adversarial manner. For the hyperparameter search we use a smaller version of ImageNet dataset (30\% of the original size) to search over the hyperparameter space. In the hyperparameter search space we search over learning rate, learning rate schedule, size of timestep and class embeddings, number of training dataset epochs. In our experiments we try two different methods of modeling the class label, a) using a class label token, b) modeling it through the cross-attention mechanism similar to the UNet architecture. We observe that configuration a) of modeling the class label as a separate token shows significant improvement in the quality of generated images. For more details related to the training details of the training setup and hyperparameters please refer to the supplementary.

\section{Results}

\noindent In this section we report results of the different architectures on the ImageNet dataset. We also compare our reported results to state-of-the-art models from ~\cite{DBLP:journals/corr/abs-2112-10752}. Our results show that the Transformer based Latent Diffusion models can get performance competitive the UNet based models. Our primary objective with this result is to show that Transformer models are a viable alternative to UNet architectures and they can potentially provide a unification paradigm across Vision and NLP architectures as the community develops new image-text multimodal models.

\subsection{Transformer v/s UNet}
\noindent Our best Transformer model attains a FID value of \textbf{14.1} which is on-par with the FID value of \textbf{13.1} for the best UNet model. For comparison we also add the best comparable models from ~\cite{DBLP:journals/corr/abs-2112-10752} in the Table \ref{table:fid_results}. Note that results from ~\cite{DBLP:journals/corr/abs-2112-10752} were trained for 1.8$\times$ number of training steps that we used for our models. This discrepancy stems from the resource intensive requirements of diffusion models.

We also report FID metric to compare the performance of standard diffusion models to the classifier free guidance models. Incorporating classifier free guidance improves the FID metric for Transformer models from 20.6 to 14.1, correspondingly it also improves the FID metric for UNet models from 19.1 to 13.1.

\begin{table}[h!]
  \begin{center}
    \label{tab:table1}
    \begin{tabular}{l|r}
      \toprule % <-- Toprule here
      \textbf{Model Architecture} & \textbf{FID $\downarrow$}\\
      \midrule % <-- Midrule here
      LDM-8-UNet & 19.1\\
      LDM-8-G-UNet & 13.1\\
      LDM-8-Transformer & 20.6\\
      LDM-8-G-Transformer & 14.1\\
      \midrule % <-- Midrule here
      LDM-8 ~\cite{DBLP:journals/corr/abs-2112-10752} & 17.4\\
      LDM-8-G ~\cite{DBLP:journals/corr/abs-2112-10752} & 8.1\\
      LDM-4 ~\cite{DBLP:journals/corr/abs-2112-10752} & 10.6\\
      LDM-4-G ~\cite{DBLP:journals/corr/abs-2112-10752} & 3.6\\
      \bottomrule % <-- Bottomrule here
    \end{tabular}
    \\
    % \addlinespace
    % \addlinespace
    \caption{Results for image synthesis on the ImageNet dataset ~\cite{DBLP:conf/cvpr/DengDSLL009}. The models are compared on the FID metric (lower is better). Our proposed model LDM-8G-Transformer achieves FID performance comparable to the LDM-8-G-UNet model. For reference we also report results from ~\cite{DBLP:journals/corr/abs-2112-10752} which were trained for 1.8$\times$ the GPU days of our models.}\label{table:fid_results}
  \end{center}
\end{table}

\subsection{Comparison to other SOTA methods, computational requirements.}

\noindent Training diffusion models is extremely resource intensive, for e.g. the LDM-4-G model from ~\cite{DBLP:journals/corr/abs-2112-10752} requires 271 V100 GPU days for training. Due to resource limitations we were able to train our UNet and Transformer models for 150GPU days each. Our LDM-8-UNet and LDM-8-G-UNet models are similar to the LDM-4 and LDM-4-G models from ~\cite{DBLP:journals/corr/abs-2112-10752}, except for the difference in the configuration of the VQGAN encoder-decoder. We used the LDM-8 VQGAN encoder-decoder as that was the best reported model when we started training our architectures, LDM-4 models give slightly better FID performance and we have added those numbers to the table \ref{table:fid_results} for comparison with our models.

\hfill

\begin{figure}[ht]
\centering
     \includegraphics[width=\linewidth]{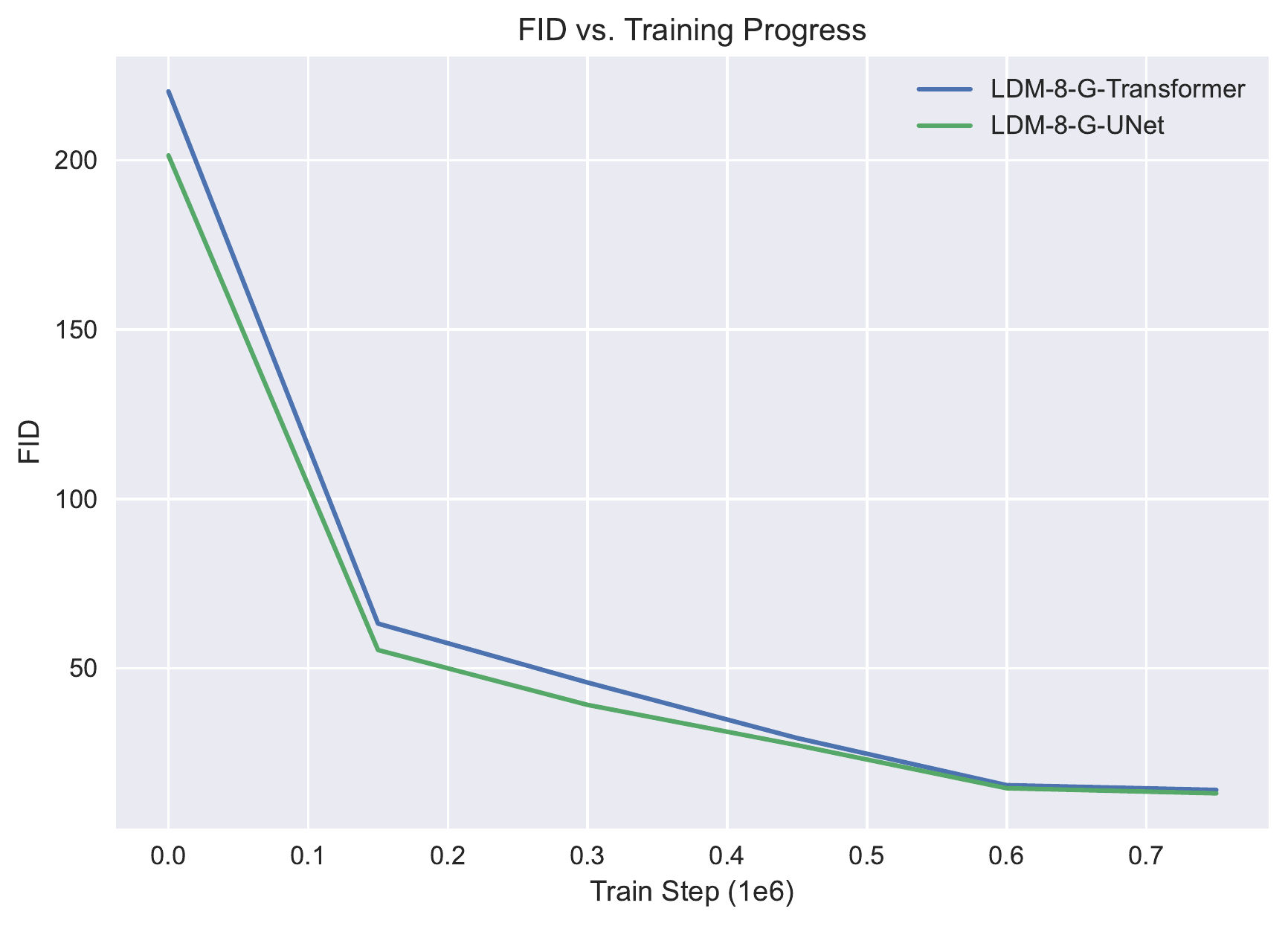}
     \caption{Progress of FID ($\downarrow$ is better) v/s training steps on the ImageNet dataset. The Transformer model learns slowly compared to the UNet model, both models converge to similar FID values at the end of training.}
     \label{fid_graph}
\end{figure}

\noindent We also plot the performance improvement in FID with number of epochs over the train dataset in fig \ref{fid_graph}. The graph shows a similar rate of improvement both for the Transformer based models relative to the UNet based models. In our experiments we observed that Transformer models are slow to learn at the beginning but eventually they converge to similar FID values as the UNet models.

\hfill

\noindent Note that our Latent Diffusion Transformer models perform on par with UNet based architectures and do not show an improvement in FID. The aim of this work is to explore usage of a Transformer based architecture for the encoder-decoder in the diffusion process, we believe that usage of Transformer simplifies the interaction between image tokens and class label. We also hope that as we move from class labels to text conditioned image generation we will observe added benefits from the usage of Transformer architecture as the architecture gives state-of-the-art performance on a wide range of NLP tasks.

\section{Conclusion}

\noindent In this work we described the design and implementation of a Latent Diffusion Transformer model, the architecture uses a native Transformer model in the latent space during the reverse diffusion process. Our experiments show that the Transformer architecture can achieve performance competitive to the UNet architecture.

We hope that this work can pave the way for a unified multimodal architecture across the vision and text domains and can also enable easy modeling of interactions between the different modalities, for example using a native Transformer removes the requirement for a dedicated cross-attention layer for modeling the text and vision features at the later stage of the UNet architecture. Finally, we believe that scaling of Transformer models with longer training times and to larger size will further improve the results.

% \clearpage

{\small
\bibliographystyle{ieee_fullname}
\bibliography{egbib}
}

\end{document}